# Integrating Natural Language Processing and Exercise Monitoring for Early Diagnosis of Metabolic Syndrome: A Deep Learning Approach

Yichen Zhao, Yuhua Wang, Xi Cheng, Junhao Fang and Yang Yang

## ABSTRACT

Metabolic syndrome (MetS) is a medication condition characterized by abdominal obesity, insulin resistance, hypertension and hyperlipidemia. It increases the risk of majority of chronic diseases, including type 2 diabetes mellitus, and affects about one quarter of the global population. Therefore, early detection and timely intervention for MetS are crucial. Standard diagnosis for MetS components requires blood tests conducted within medical institutions. However, it is frequently underestimated, leading to unmet need for care for MetS population. This study aims to use the least physiological data and free texts about exercises related activities, which are obtained easily in daily life, to diagnosis MetS. We collected the data from 40 volunteers in a nursing home and used data augmentation to reduce the imbalance. We propose a deep learning framework for classifying MetS that integrates natural language processing (NLP) and exercise monitoring. The results showed that the best model reported a high positive result (AUROC=0.806 and REC=76.3%) through 3-fold cross-validation. Feature importance analysis revealed that text and minimum heart rate on a daily basis contribute the most in the classification of MetS. This study demonstrates the potential application of data that are easily measurable in daily life for the early diagnosis of MetS, which could contribute to reducing the cost of screening and management for MetS population.

## 1 Introduction

Metabolic syndrome (MetS) is a medication condition characterized by abdominal obesity, hypertension, dyslipidemia, and elevated fasting glucose levels[1]. It has been proven to increase the risk of various diseases, such as type 2 diabetes mellitus, cardiovascular diseases, chronic kidney disease, cancer, and also affects patients' physical mobility. Over the past 2 decades, the number of individuals with MetS worldwide has surged dramatically, becoming a serious public health issue[2]. A meta-analysis of epidemiological studies on MetS showed that the prevalence of MetS in China is 21.90%[3]. Early warning of the risk of MetS and timely intervention and prevention are extremely important.

Currently, there are certain difficulties in the popularization of early diagnosis and intervention for MetS. Firstly, MetS-related indicators such as triglycerides, high-density lipoprotein cholesterol, and fasting blood glucose generally require blood tests conducted within professional medical institutions. At the same time, the confirmation of MetS requires metabolic experts in the medical structure, and the missed diagnosis rate is relatively high among outpatients in other specialty departments. Secondly, there is still a lack of effective clinical management measures for MetS patients, especially those who also suffer from other diseases.

Lifestyle is the main intervention for MetS. Most MetS patients exhibit a decline in physical mobility, which involves complex pathogenesis such as insulin resistance, endothelial dysfunction, oxidative stress, inflammation, and changes in lipid metabolism[4]. Studies have found that even after correcting factors such as BMI, the upper limit of exercise in men with MetS is significantly lower than that in men without MetS, and among patients with MetS, the proportion with exercise habits is relatively small[5]. On the other hand, exercise can effectively regulate the negative impact brought by MetS. Studies have pointed out that individuals with MetS have seen significant improvements in their condition after engaging in more daily physical activities[6], and planned exercise (including walking and other daily activities) can also benefit MetS patients with risk factors[7]. Therefore, efficiently monitoring and precise assessment of daily exercise would address the unmet needs for care of MetS population. Specifically, monitoring daily exercise can be used to assess the risk of individuals developing MetS. Meanwhile, precise assessment the association between daily exercise and MetS can support and improve self-management for MetS population.

Wearable devices, through tracking a variety of health-related variables, are one of the main ways to monitor and assess exercise [8]. However, due to the large differences in acceptance of wearable devices among different populations, poor compliance, difficulties in long-term wearing of wearable devices, limited types of measured data (often estimating other data based on a small amount of measured data), unsatisfied data quality, difficulties in data fusion from multiple devices, and inaccurate data monitoring during exercise[9], it is not enough to assist in the diagnosis of MetS by analyzing data provided solely by wearable devices.

Therefore, we introduced Natural Language Processing (NLP), which has proven to be powerful in semantic understanding and medical information summarization capabilities[10]. We first extracted relevant exercise textual information from individuals at risk of MetS regarding their own exercise descriptions, and then integrated the least physiological data about exercise using wearable devices, to learn latent exercise habits and effects, with the expectation of assisting in the diagnosis of MetS.

To our knowledge, this is the first study to integrate NLP and physiology measurements to learn latent exercise capacity and habits, thereby assisting in the early diagnosis of MetS. We have validated this hypothesis by collecting a dataset and establishing a model, which has yielded significant results. It not only provides a valuable supplement to existing models for predicting and managing the risk of MetS, but also holds the potential to be integrated into online tools, promoting the widespread adoption of disease screening and health awareness, and serving a larger group of high-risk population.

## 2 Method

### 2.1 Data Collection and Preprocessing

We recruited 40 volunteers from a nursing home, and collected their most recent physical examination data. According to the diagnostic criteria for MetS recommended by the Diabetes Society of the Chinese Medical Association[11], individuals with overweight and/or obesity (BMI≥25.0 kg/m²), high blood sugar (FPG≥6.1 mmol/L (110 mg/dl) and/or 2hPG≥7.8 mmol/L (140 mg/dl)), and/or diagnosed and treated diabetes, hypertension (SBP/DBP ≥ 140/90 mmHg), and/or diagnosed and treated hypertension, and dyslipidemia (fasting blood TG≥1.7 mmol/L (110 mg/dl) and/or fasting blood HDL-C < 0.9 mmol/L (35 mg/dl) for males, < 1.0 mmol/L (39 mg/dl) for females) were selected. Samples meeting 3 or more of the 4 criteria were divided into 2 groups: those with MetS and those without. After grouping, 8 individuals were classified as having MetS, and 32 as not having MetS.

Data collection was conducted on a daily basis, including four physiological characteristics (daily minimum heart rate, daily maximum heart rate, daily average oxygen saturation, daily step count) and one textual feature (subjective text on exercise experience). In line with the research objectives of this project, the physiological characteristics selected were easily accessible indicators, all of which were measured using the Huawei Band 8 (https://consumer.huawei.com/cn/wearables/band8/). During the pre-experimental phase, a questionnaire was used to guide the volunteers to familiarize themselves with the key points of language expression and to help extract critical information. The questions included the type and duration of exercise or household chores performed by the volunteers that day, their physical sensations during the process (such as shortness of breath, slight sweating, severe panting, chest tightness, or hypoglycemia), and the impact of weather and other factors on the activity. Subsequently, in the formal experiment, the provided text content mainly includes the following aspects: the type and duration of exercise or household chores, physical sensations, and influencing factors. To ensure the quality of the data, we provided a relevant text template that includes the above content. At the same time, based on the feedback from volunteers and the principle of minimal interference with daily life, we allowed volunteers to elaborate and supplement based on the actual text framework. We conducted two phases of data collection for the aforementioned group, with each phase lasting 14 days. Considering the imbalance of the sample, an additional phase of data collection lasting 10 days was conducted for the group with MetS to increase the volume of data. The data collection process is shown in **Appendix 1**.

Subsequently, the collected physiological data and textual data were subjected to data cleaning, including the exclusion or imputation of missing and outlier values. For data missing subjective exercise experience text, deletion was performed due to the inability to reasonably impute; for missing physiological data, the average value of that volunteer's data for the item was used as a substitute; for outlier data, such as erroneous physiological data caused by incorrect use of the smart bracelet by the volunteer, manual correction or deletion was carried out depending on whether the correct value could be found; for individual volunteer data of poor quality, such as excessive missing data or poor voice text quality, all data for that volunteer were deleted as a whole (2 non-MetS volunteers' data were completely removed for this reason). After preprocessing, the group with MetS included 8 individuals with 290 sets of data, and the group without MetS included 30 individuals with 649 sets of data.

### 2.2 Model Architecture

Based on the physiological data as well as the exercise-related textual data, we have proposed a classification framework for MetS, as shown in **Figure 1**.

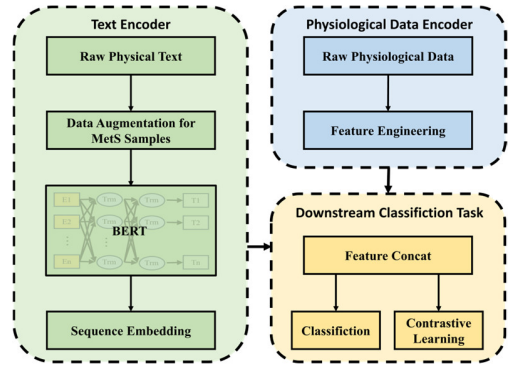

**Figure 1: Text-physiological modality fusion for MetS classification framework. The framework is divided into three parts. The Text Encoder transforms textual information into Sequence Embedding. The Physiology Data Encoder section processes raw physiological data through feature engineering to become model inputs. The Downstream Classification Task section trains the classification model with the fused inputs obtained by the text and physiology data encoder.**

Initially, based on different fusion strategies, we proposed 3 types of model architectures, as depicted in **Figure 2**. We evaluated the performance of each model architecture by the optimal Area Under the Receiver Operating Characteristic Curve (AUROC) after convergence. Subsequently, we conducted a grid search for the optimal combination of hyperparameters involved in the models, which include the neuron sizes of the 2 hidden layers termed Reduced Dimension and Hidden Layer Dimension, as well as the probability p of dropout. During the hyperparameter search process, we assessed model performance based on the maximum AUROC value after model convergence. By comparing the performance on the validation set, we determined that the optimal settings were a Reduced Dimension of 3, a Hidden Layer Dimension of 32, and a dropout probability p of 0.3, which yielded the best results on the validation set.

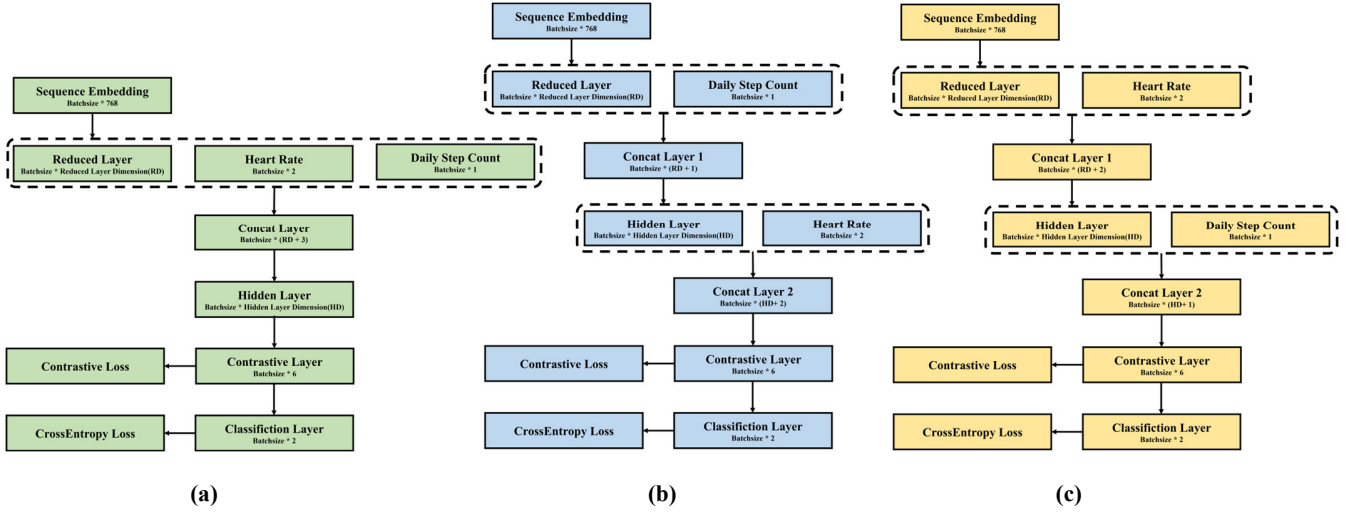

Figure 2: Three model architectures. (a) THSCL Model: daily text, daily heart rate and daily step count are concatenated simultaneously; (b) TS_HCL Model: daily text and daily step count are concatenated first, followed by daily heart rate; (c) TH_SCL Model: daily text and daily heart rate are concatenated first, followed by daily step count. For three architectures above, contrastive loss is added.

Additionally, we designed a Baseline Model to assess the advantages of the proposed model architecture. In the Baseline Model, the output of the BERT-styled Transformer is directly concatenated with the physiological data and then passed through a Multilayer Perceptron (MLP) for binary classification. In Baseline Model, contrastive loss is not introduced. In this model, dropout is utilized in the hidden layers to mitigate overfitting as much as possible, with a learning rate set to $10^{-4}$ and the optimizer configured as AdamW.

## 2.3 Data Augmentation and Model Training

Initially, to eliminate redundant information that may affect decision-making, we conducted feature selection on the preprocessed data. We randomly sampled from the original physiological data of both the MetS group and the non-MetS group, and excluded features that did not exhibit significant differences between the 2 groups through significance testing. Subsequently, we performed data augmentation. For MetS group, we enhanced the higher-quality texts by manually removing irrelevant information to generate new textual data. Additionally, we augmented the exercise texts from all individuals in the MetS group by translating them into another language and then back into Chinese, thereby increasing the diversity of the data. The purpose of data augmentation was to enhance the diversity of the MetS data and to balance the data volume with the non-MetS group. After augmentation, the MetS group retained 656 sets of data. Thereafter, all physiological data were normalized to serve as inputs for the model.

Following data augmentation, the dataset was divided. First, the dataset was partitioned into training and validation sets to adjust the model architecture and hyperparameters. To ensure a balanced representation of individuals with and without MetS in both the training and validation sets, we randomly selected data from 2 individuals in the MetS group and, based on the quantity of data from the selected individuals, randomly sampled a similar amount of data from the non-MetS group for the validation set, with the remaining data used for training. Each volunteer's data from the 38 individuals was included in either the training set or the validation set, ensuring the independence of the 2 sets. Secondly, to comprehensively assess the optimal model performance and prevent the poor selection of training and testing sets from failing to reflect the performance of the best model, we randomly selected 25% of the data as an independent test set and evenly divided the remaining 75% of the data into 3 folds for 3-fold cross-validation. In each evaluation, we chose two folds as the training set and one fold as the validation set, selected the optimal model on the validation set and then assessed the model's generalizability on the test set.

The model was trained using a combined loss function of cross-entropy loss and contrastive loss.

$$L = \alpha \cdot L_{CE} + (1-\alpha) \cdot L_{CON} \quad (1)$$

$$L_{CE} = -(y \log(\hat{y}) + (1-y) \log(1-\hat{y})) \quad (2)$$

$$L_{CON} = [y_i = y_j]||\theta(x_i) - \theta(x_j)||_2^2 + [y_i \neq y_j]\max(0, \varepsilon - ||\theta(x_i) - \theta(x_j)||_2^2) \quad (3)$$

In the model training, we introduce a combined loss function where $\alpha$ is set to 0.7 to balance 2 loss values to an approximately equal value and $\varepsilon$ is set to 0.5. $y$ represents the true classification labels, and $\hat{y}$ denotes the predicted probability distribution after softmax processing. $x_i$ is the input, and $\theta(x_i)$ is the result of the sample $x_i$ processed by the model with parameters $\theta$. We incorporate contrastive learning into the model training to enable the model to discern the subtle differences between MetS positive and negative samples, thereby enhancing the model's classification performance.

## 2.4 Model Evaluation

Considering the practical requirements of medical classification models, we assess the performance of the model using accuracy, recall, precision, F1 Score, and AUROC (Area Under the Receiver Operating Characteristic Curve).

$$ACC = \frac{TP + TN}{TP + TN + FP + FN} \quad (4)$$

$$REC = \frac{TP}{TP + FN} \quad (5)$$

$$PRE = \frac{TP}{TP + FP} \quad (6)$$

$$F1\,Score = \frac{2 \times PRE \times REC}{PRE + REC} \quad (7)$$

In the context of medical classification models, we evaluate the model's performance using true positives (TP), which represent the number of correctly classified individuals with MetS; true negatives (TN), representing the number of correctly classified individuals without MetS; false positives (FP), which are the number of individuals without MetS incorrectly predicted as having it; and false negatives (FN), the number of individuals with MetS incorrectly predicted as not having it.

At the same time, we consider the impact of the imbalance between positive (MetS) and negative (non-MetS) samples on the model's performance. Since the prevalence of the vast majority of diseases in the population is a relatively low proportion, this is a problem that all current disease prediction model designs have to face. We control the degree of data augmentation to achieve different ratios of MetS and non-MetS samples, and compare the performance of the models trained by them, thereby reflecting the relationship between the model's predictive performance and the degree of imbalance to some extent.

## 2.5 Feature Importance Analysis

To enhance the interpretability of the model and identify features associated with MetS, we employ the Permutation Feature Importance (PFI) algorithm[11] to quantify the importance of each feature. The process involves four main steps:

1. Input the original test data into the model to obtain the initial performance assessment;
2. For each feature, randomly shuffle the feature vector within the input data to create modified test data;
3. Input the modified test data into the model to obtain a new performance assessment, and define the importance of the feature using the following formula.

$$P = P_r - \frac{1}{k}\sum_{i=1}^{k} P_i \quad (8)$$

$P_r$ represents the performance assessment of the model when the original test data is inputted, and $P_i$ represents the performance assessment of the model after the $i$th permutation of the feature. $k$ denotes the number of times the feature is randomly permuted, which we set to 50.

4. Execute the above steps for each feature; the higher the $P$ value, the greater the importance of the feature. To comprehensively evaluate the model's classification performance, we adopt the AUROC as the performance assessment metric to determine the importance of each feature.

Additionally, we employed LIME[12] for token-level analysis of textual data. LIME is an algorithm designed to explain the local prediction behavior of complex models, with the core concept of constructing a simple, interpretable model to simulate the decision-making of the complex model in the vicinity of specific data points. Based on the LIME algorithm, we are able to more clearly identify which parts of the textual data influence the model's decision-making. We will provide a specific case in Section 3.3 to assist in illustrating the analysis results.

## 3 Result

We analyzed the physical examination data collected from 40 elderly individuals at a nursing home. **Table 1** describes the statistical characteristics of the study population. Compared to the non-MetS group, there were extremely significant differences in waist circumference, body mass index, triglycerides, and fasting blood glucose (p<0.01).

**Table 1: Statistical Characteristics of Volunteers with MetS and non-MetS**

| Characteristics | Mets | Non-Mets |
| --- | --- | --- |
| N | 8 | 32 |
| Sex (men%) | 12.50 | 10.00 |
| Age | 74.88 | 72.19 |
| Height (cm) | 157.44 | 160.13 |
| Body Mass Index (BMI) | 28.21 | 21.15 |
| Waist Circumference (cm) | 94.83 | 80.28 |
| Systolic Blood Pressure (mmHg) | 143.88 | 133.72 |
| Diastolic Blood Pressure (mmHg) | 79.63 | 78.41 |
| High-Density Lipoprotein (mmol/L) | 1.51 | 1.47 |
| Triglycerides (mmol/L) | 2.09 | 1.36 |
| Fasting Blood Glucose (mmol/L) | 6.99 | 5.42 |

### 3.1 Comparison of Physiological Data for Assisting in the Diagnosis of MetS Risk

We performed a random sampling on the original data that had been cleaned, with missing values and outliers excluded or imputed. We extracted data from eight individuals without MetS (170 sets) and compared it using a t-test with the original data from eight individuals with MetS (290 sets). The significantly different features are described in **Appendix 2**. The results indicate that daily minimum heart rate, maximum heart rate, and step count exhibit significant differences between the groups with and without MetS ($p<0.01$), showing potential in assisting the diagnosis of MetS.

**Table 2: Assessment Results of Baseline, TS_HCL and TS_H**

| Model | Datasets | ACC(%) | PRE(%) | REC(%) | F1(%) | AUROC (%) |
|---|---|---|---|---|---|---|
| Baseline | Train (Fold2, 3), Val (Fold1) | 68.8 | 70.1 | 63.5 | 66.7 | 76.0 |
| | Train (Fold1, 3), Val (Fold2) | **72.3** | **73.1** | 68.8 | 70.9 | 80.2 |
| | Train (Fold1, 2), Val (Fold3) | 71.4 | **71.3** | 70.0 | 70.6 | 79.7 |
| | Average (Std) | 70.8 (1.47) | **71.5 (1.24)** | 67.5 (2.81) | 69.4 (1.94) | 78.6 (1.87) |
| TS_H | Train (Fold2, 3), Val (Fold1) | 69.9 | **72.0** | 63.5 | 67.5 | 76.8 |
| | Train (Fold1, 3), Val (Fold2) | 71.7 | 68.6 | 78.2 | **73.1** | **83.5** |
| | Train (Fold1, 2), Val (Fold3) | 71.4 | 70.5 | 71.8 | 71.1 | 81.2 |
| | Average (Std) | **71.0 (0.76)** | 70.4 (1.41) | 71.2 (6.02) | 70.6 (2.31) | 80.5 (2.79) |
| TS_HCL | Train (Fold2, 3), Val (Fold1) | **70.5** | 71.3 | **67.1** | 69.1 | 77.0 |
| | Train (Fold1, 3), Val (Fold2) | 69.1 | 64.3 | **83.5** | 72.6 | 82.3 |
| | Train (Fold1, 2), Val (Fold3) | **71.7** | 68.6 | **78.2** | 73.1 | 82.2 |
| | Average (Std) | 70.4 (1.06) | 68.0 (2.88) | **76.3 (6.87)** | **71.6 (1.78)** | **80.6 (2.53)** |

### 3.2 Model Evaluation Combined with NLP

For the three proposed model architectures and hyperparameters, we employed grid search to identify the optimal combination of model architecture and hyperparameters, assessing the model performance based on the highest AUROC. The best model negative samples, TS_HCL incorporates the contrastive loss, resulting in the best classification performance compared to the baseline model and TS_H (as shown in **Table 2**). Overall, the TS_HCL performed the best AUROC, followed by TS_H. TS_HCL improved the AUROC by 2.0% comparing to the Baseline Model, and by 0.1% compared to TS_H. Besides, TS_HCL performed the best recall, which improved the REC by 8.8% comparing to the Baseline Model, and by 5.1% comparing to TS_H. This indicates that the TS_HCL model can better reduce the false-negative rate. The improvement of the AUROC and the REC illustrates the superiority of using contrastive loss for this task.

Further investigation into the impact of the imbalance between positive (MetS) and negative (non-MetS) samples on the model's performance is shown in **Figure 3**. We alter the proportion of MetS in the training set and compare the model's performance on the test set using AUROC and ACC, as is shown in Figure 3. When the proportion of MetS is low (between 30% and 35%), the model is unable to correctly classify MetS and non-MetS; it tends to predict all training data as non-MetS. As the proportion of MetS increases, the model's performance increases too, and the model converges more rapidly. The model's performance reaches its peak when the positive and negative samples are balanced.

### 3.3 Feature Importance Ranking

To ascertain the contribution of textual and physiological data to the model, we employed an evaluation method based on the Permutation Feature Importance (PFI) algorithm to rank the features, as depicted in **Appendix 3**. The results indicate that when textual data is compromised, the model's assessment performance declines by approximately 21.2%, which is the greatest decrease; when the daily minimum heart rate data is compromised, the model's assessment performance declines by about 4.2%, which is the largest decrease among the physiological data when compromised, but significantly less than the impact of the textual data's failure on the model. In contrast, the decline is not significant after shuffling the daily maximum heart rate and daily step count, with reductions of only 1.2% and 0.8%, respectively.

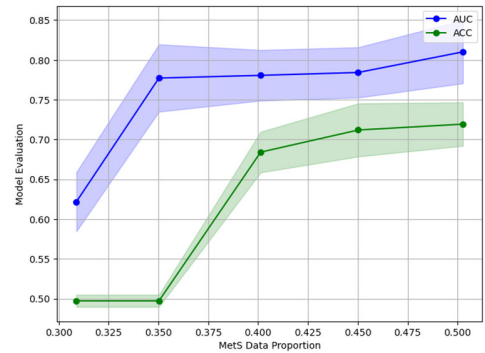

**Figure 3: Influence of MetS Data Proportion on Model Effect. The horizontal axis represents the proportion of MetS data in the total data volume, and the vertical axis represents the average model performance evaluation (ACC and AUROC) across four folds. The figure also shows the 95% confidence interval.**

Thus, it can be concluded that the decision-making process of our model primarily relies on textual data. Furthermore, among the physiological data, the model evaluation mainly depends on the data for daily minimum heart rate.

To delve deeper into which parts of the textual data affected the model's decision-making, we utilized LIME algorithm for token-level analysis. **Figure 4** presents the analysis results of the LIME algorithm on the textual data from a day of a MetS patient. This volunteer was correctly classified by the model as a MetS patient, and the primary aspect influencing the model's decision was the negative feelings expressed by this volunteer during or after exercise.

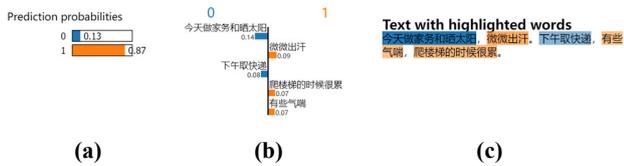

**Figure 4: Token-level analysis using LIME on a case study. (a) The classification outcome of the model; (b) The contribution of different parts of the textual data to the model's decision-making; (c) Text with highlighted words. The textual data is in Chinese, and for ease of understanding, we translated it into English: "I did household chores and sunbathed today, breaking into a slight sweat; in the afternoon I fetched the courier, felt a bit out of breath, and was very tired when climbing the stairs."**

## 4. Discussion

Utilizing the high patient compliance, simple interaction, and user-friendly advantages of NLP combined with exercise monitoring, we have constructed a machine learning model to predict the risk of MetS, in order to facilitate the diagnosis and enhance the probability of early diagnosis of MetS in the population. We designed 3 model architectures and selected the optimal model and parameters in the validation set. On the test set, the model reported a high positive result through 3-fold cross-validation (AUROC=0.806). In addition, our model TS_HCL exhibited the highest Recall (REC=76.3%) compared to TS_H and the Baseline Model, suggesting that our model can avoid type II error more effectively, preventing patients suffering from the disease from not being treated in a timely manner due to model prediction errors. By ranking feature importance, we found that the decision process of this model is most dependent on text, indicating that the application of NLP to exercise monitoring can assist in the diagnosis of MetS. Consistent with current research on MetS, we concluded that exercise habits and capacity can to some extent reflect a person's risk of having MetS. Utilizing NLP to process personal descriptions of exercise habits and capacity can serve as a new method for early diagnosis in the management of MetS.

NLP can be widely applied to the processing of sequential data. Early diagnosis of some diseases often becomes complex due to various factors in diagnostic indicators, and standard tests usually require a lot of experience and effort. Our study innovatively approaches the assessment of an individual's risk of MetS from the perspective of exercise capacity and ability to judge exercise habits, proposing the use of NLP to monitor text provided by volunteers describing their exercise capacity and habits, and has developed and validated the model accordingly. Since exercise texts often contain other factors in addition to the main research factors, such as weather and diet, which also affect the subjective feeling of exercise, these factors may also be reflected in the exercise texts. NLP can utilize this co-occurring information simultaneously, identify confounding information, and enhance the effectiveness and accuracy of the analysis of individual exercise habits and capacity, thereby achieving the accuracy and reliability of early diagnosis of MetS.

Considering that the NLP-based exercise monitoring early diagnosis model for MetS is non-invasive, suitable for daily exercise health management, and can make up for the subjective and objective factors ignored by wearable device monitoring of exercise indicators, we believe that it can greatly improve user compliance and realize a new way of exercise early diagnosis for MetS. On the one hand, it is expected to promote the popularization of early diagnosis of MetS, and on the other hand, it is expected to address the issues of low diagnostic efficiency and high cost of MetS.

Previous studies have also done a lot of work on the early diagnosis of MetS, mainly involving predictive models developed based on other machine learning methods and the biomarkers detection. Since 2015, most of the non-invasive predictive model studies on MetS have been about lifestyle-related characteristics and anthropometric characteristics[13, 14, 15, 16, 17, 18, 19], but these models mainly use anthropometric characteristics as the main features, and the role of lifestyle characteristics is not significant[20, 21]. Our early diagnosis model starts from monitoring lifestyle characteristics, and its significant effect has been verified, which to some extent fills the gap in this field.

The flexibility of deep learning poses a challenge in selecting the most appropriate neural network architectures for a specific dataset to achieve optimal model performance. Based on the information contained in the three modalities, we initially designed three model architectures (as shown in Figure 2) and then used a grid search to determine the best combination of hyperparameters. By evaluating the validation set, the TS_HCL (Figure 2(b)) performed the best. This model architecture first encodes daily text, followed by fusion with the daily step count, then fuses the features of the heart rate, and finally feeds them into the classification head. The two-step fusion implies the correlation among the three modalities with the designed loss function. Intuitively, we grouped the daily minimum and maximum heart rates together to allow the model to synthesize heart rate information; we grouped the daily text and daily step count together to enable the model to more comprehensively assess the individual's exercise duration and intensity. Furthermore, contrastively learning enhanced TS_H by introducing the contrastive loss, which has been proven to be effective for classification in many other domains. **Table 2** demonstrates the model performance improvement with our designed model architecture and the introduction of contrastive loss. The average AUROC of TS_H is improved by 1.9% compared to the Baseline Model, while the average AUROC of TS_HCL is the highest among the three models, with an improvement of 2.0% compared to the Baseline Model. Meanwhile, TS_HCL performed the best recall. The REC of TS_HCL is improved by 8.8% compared to the Baseline Model, and by 5.1% compared to TS_H.

When designing disease prediction models, the prevalence of the disease (i.e., class imbalance) significantly affects the model's performance. Using proper sample balancing techniques (e.g., data augmentation) can effectively improve the model's performance and generalizability, and ensure robust model when faced with real-world imbalanced disease data. It is worth noting that most existing data augmentation techniques tend to generate new data that are correlated with the original data to a certain extent, resulting in the problem of homogenization, which may negatively impact on the deployment of the model in real world. It is crucial to have the developed model to be tested under different imbalance ratios, and select the model that meets the criteria for specific application.

To enhance the interpretability of the model, when we performed feature importance ranking on the model, we found that the daily minimum heart rate also showed the second-highest weight in the model's decision-making process after textual data. Combined with the results of the significant comparison of the collected physiological data, there is indeed a very significant difference in daily minimum heart rate between the group with MetS and the non-MetS, and the coefficient of variation (CV=9.1%) of the MetS group is lower than that of the non-MetS group (CV=15.5%). Studies by Gupta and Schroeder[22, 23] have shown that low heart rate variability is related to cardiovascular diseases, and combined with the fact that MetS is also related to a series of cardiovascular risk factors, we believe that daily minimum heart rate may be an important feature for prediction, providing a guiding direction for the subsequent improvement of the model.

To our knowledge, this is the first study to propose the use of NLP to monitor exercise capacity and habits, thereby assisting in the diagnosis of MetS. We have validated this hypothesis by collecting a dataset and establishing a model, which has achieved significant results. It not only provides a valuable supplement to existing models for predicting and managing the risk of MetS but also has the potential to be integrated into online tools to serve a larger group of potential patients.

On the technical level, NLP addresses the limitations of monitoring capabilities affected by subjective and objective factors in wearable devices. By further collecting volunteers' exercise texts (including exercise time, exercise intensity, exercise type, physical condition, etc), it supplements the subjective and objective factors ignored by wearable devices in monitoring exercise indicators. On the economic level, the early diagnosis method through NLP can improve the current situation of high screening costs for MetS, and the non-invasive method also makes individuals more willing to undergo long-term health supervision, promoting the popularization of disease screening and health awareness, and vigorously developing the construction of healthy.

A limitation of this study is that it still lacks the validation in generalizing the predictive model to a broader population. MetS is influenced by various factors such as race and environment. Although our method has been verified and tested in the current cohort, it is limited to a subset of the elderly population in a specific city in China, and its performance in other regions and ethnicities remains uncertain. The sample lacks diversity and requires assessment through more external validation datasets, especially those involving other ethnic populations.

Secondly, during the experimental process, we found that the richness of the textual data provided by different volunteers varies due to differences in educational levels, which indirectly affects the accuracy of the model's judgment. During the formal experimental phase, we mitigated these differences and enhanced text quality by providing a text template.

Our assessment of the most influential textual data may also have another interpretation. Due to the utilization of textual data as a multimodal model with the highest dimensionality among data modalities, it occupies a significant proportion in the model's predictions, while other data modalities may be overlooked. However, this also reflects that the analysis of textual data has certain effectiveness in the prediction of MetS.

## 5. Conclusion

Our study primarily utilized NLP to process the textual data collected from volunteers, developing a predictive model for MetS aimed at identifying potential individuals with MetS from the regular textual data provided by groups with existing risk factors. The model, after 3-fold cross-validation, reported a high positive result (AUROC=0.806 and REC=76.3%), demonstrating its practicality and reliability in future predictions of MetS. This will aid in promoting the advancement of digital and precision medicine in the early diagnosis and intervention among groups with potential risk factors for MetS, reducing the costs associated with widespread screening for MetS, thereby improving the disease management of MetS.